# An optical diffractive deep neural network with multiple frequency-channels

Yingshi Chen[1], Jinfeng Zhu[1]

[1] Institute of Electromagnetics and Acoustics, and Department of Electronic Science, Xiamen University, Xiamen 361005, China
E-mail: gsp@grusoft.com

**Abstract**

Diffractive deep neural network (DNNet) is a novel machine learning framework on the modulation of optical transmission. Diffractive network would get predictions at the speed of light. It's pure passive architecture, no additional power consumption. We improved the accuracy of diffractive network with optical waves at different frequency. Each layers have multiple frequency-channels (optical distributions at different frequency). These channels are merged at the output plane to get final output. The experiment in the fasion-MNIST and EMNIST datasets showed multiple frequency-channels would increase the accuracy a lot. We also give detailed analysis to show the difference between DNNet and MLP. The modulation process in DNNet is actually optical activation function. We develop an open source package ONNet. The source codes are available at https://github.com/closest-git/ONNet.

**1. Introduction**

Diffractive deep neural network(DNNet)[1,2] is a novel machine learning framework on the physical principles of optics, which is still in its infancy and shows great potential. DNNet tries to find optimal modulation parameters to change the phase, amplitude or other physical variable of optical wave propagation. So in the final output plane, the optical distribution has special pattern which is the indicator of object's class or value. DNNet tries to extract features from the wave distribution in the three dimensional space, which is fundamentally different from the images, audios or videos. The learning of wave propagation has hardly been studied in machine learning. This is a new subject. DNNet opens new doors for the machine learning.

If must have to pick a reference, the most similar tool in the machine learning is multiple layer perceptron (MLP) [10]. [1] gives some comparison between the structure of diffractive neural networks and electronic networks, which is actually MLP. As we showed in section 2, there are still many big differences. Compared to more complex deep convolution neural networks (CNN), DNNet lacks more key features: 1) No convolution layers. 2) No batch normalization or other normalization module. 3) No pooling layers. There are some ways to implement more complex optical neural networks, which have optical normalization layers and optical pooling layers (We will discuss this in the following papers). But the first point is the key difference. There's no sign how to add some optical kernels to learn the various details and hierarchical features of input images. So learning ability of DNNet is much weak than CNN, which is also verified by our testing results. Even in the tiny fashion-MNIST and EMNIST [19] datasets, the performance is not satisfactory. So there is long way for the DNNet to catch up with his cousin, deep MLP or deep CNN.

In this article, we present a new diffractive network with multiple frequency-channels (MF_DNet). We define frequency channel as the optical distribution corresponding to a certain frequency at each layer. As figure 1 showes, we use optical waves at different frequency to get more frequency-channels. Then we merge these channels to get final result at the output layer. Experiments show that this method improves the accuracy. The $D^2$NNet in [1] is the simplest MF_DNet. Or MF_DNet with only one channel.

For the simulation of optical neural networks, we developed ONNet. It's an open-source Python/C++ package and the codes are available at https://github.com/closest-git/ONNet. ONNet implements diffractive layer on the module of pytorch [15], which is one of the best deep learning frameworks. With the powerful automatic differentiation [17] in pytorch, it's easy to get the gradient for all operations in the complex diffractive process. For example, the fast Fourier transform (FFT) and inverse FFT operator, which is the key operator for the numerical simulation of diffractive process. The main disadvantage of pytorch is the lack of support for complex tensors. ONNet implemented various complex-number tensor operations. As mentioned above, DNNet is not CNN, so ONNet has not use any convolution layers in the current release.

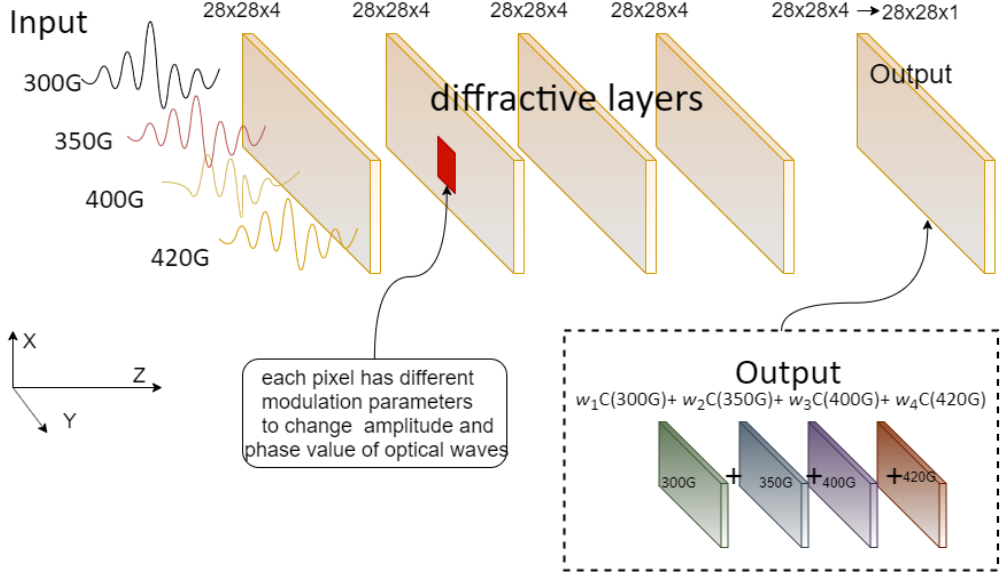

Figure 1 MF_DNet - Diffractive deep neural network with multiple frequecncy-channels

So the main contributions of this paper are as follows. 1) A detailed and in-depth analysis of DNNet's structure. We compared the similarity and difference between DNNet and MLP. The modulation process in DNNet is actually optical activation function. 2) We present MF_DNet, which improve DNNet with multi-frequency channels. The more channels, the higher accuracy. 3) An open source package ONNet that will bring a lot of help to the simulation of optical neural networks.

## 2. Analysis of DNNet

For the wave from $i$th pixel in the $l$th layer, formula (1) defines its distribution in the three dimensional space.

$$\omega_{i,p}^l = \frac{p-p_i}{r^2}\left(\frac{1}{2\pi r} + \frac{1}{j\lambda}\right) exp\left(\frac{j2\pi r}{\lambda}\right) \qquad (1)$$

Where $r$ is distance between the point $p$ and $p_i$; $j = \sqrt{-1}$; $\lambda$ is the wavelength.

For the detail proof of formula (1), please see [1]. DNNet uses stacked layers to adjust the propagation of optical waves. As figure 1 showes, each pixel in each diffractive layer(including input and oupt layer) has different modulation parameters. They will change the amplitude and phase value of optical waves. So in the final output layer , there are special optical pattern. Some regions are more brighter than other regions. By adjusting these parameters, we can change the position of brighter regions. To use these patterns as the indicator of class number, we must solve a very difficult optimazation probem. Machine learnig is a poweful tool for this problem. And experiments verified that we do find possible solution with acceptable accuracy. This stacked multilayers are similar to the structure of MLP, but there are still many big differences.

To compare classical multiple layer perceptron (MLP) and DNNet, we first give some concise definition of MLP. Let complex function $f(x;\theta)$ is the target function parameterized by $\theta$ in complex space $\mathbb{C}$. Given training data D = $\{(x_i, y_i), i = 1,2,\cdots,M\}$ where x denotes the input and y the class number. A MLP try to learn the function $f(x;\theta)$ by neural network usually consists of stacked layers. The learned model should fit the training data well and have good generalization in the testing data. Let L denotes the total number of layers and $l \in \{1,2,\cdots,L\}$, formula (2) defines the transformation in the classical MLP, which consists of a linear mapping $Z^l = (W^l)^T h^{l-1} + b^l$ and followed by an element-wise nonlinearity activation function: $h^l = \emptyset(z^l)$.

$$\begin{cases} h^l = \emptyset(z^l) \\ z_i^l = \sum_k w_{k,i}^{l-1} h_k^{l-1} + b^l \end{cases} \qquad (2)$$

$w_{k,i}^{l-1}, b^l \in \mathbb{C}^n$ are learnable parameters

Formula (3) defines transformation in the DNNet [1, 2].

$$\begin{cases} h_i^l = t_i^l z_i^l \\ z_i^l = \sum_k \omega_{k,i}^{l-1} h_k^{l-1} \\ t_i^l = a_i^l exp(j\phi_i^l) \end{cases} \quad (3)$$

where

$\omega_{i,p}^l = \frac{p-p_i}{r^2}\left(\frac{1}{2\pi r} + \frac{1}{j\lambda}\right) exp\left(\frac{j2\pi r}{\lambda}\right)$ are constants for the wave propagation

$a_i^l$(amplitude), $\phi_i^l$ (the pahse value ) are learnable parameters

Compare formula (2) and formula (3), we could see three differences.

1) The difference meaning of transfer matrix $W^l$. In the case of DNNet, $\omega_{k,i}^{l-1}$ is not learnable parameters, just some constants from the numerical simulation of wave propagation. In fact, ONNnet package calculate these parameters before the training process. And reuse these values to reduce computation time. So for each hidden layer with N neurons, MLP has N × N parameters, while DDnet has only 2N parameters. In some implementions, $a_i^l$ is always 1 and only $\phi_i^l$ are adjustable parameters, so only N parameters in this case.

2) DNNet has special activation function. That is, $h_i^l = t_i^l z_i^l$ is an optical activation defined in the complex space. [1] declares $t_i^l$ is something like bias item. But we think that it just special activation functions. In fact, we can add extra bias item as discussed in the next point.

3) No bias item $b^l$. We have tried to add bias item $b^l$ in the DNNet, that is,
$$h_i^l = t_i^l z_i^l + b^l.$$
But no significant change in the accuracy. We would give detailed analysis in the following paper.

## 3. Diffractive network with multiple frequency-channels

For each diffractive layers, there are different optical distribution at different frequency. We call these optical distributions as **frequency-channels**, just like different channels in the CNN. It's well knowned that deep CNN would get higher accuracy with more channels. So we get similar results in the DNNet, higher accuracy with more frequency-channels.

At the output layer, we get frequency-channel at different frequncy. Then we merge these channels to get final result with weighting coefficient. The loss function is defined in formula (4).

$$\text{Loss} = loss\left(\sum w_f \times channel\right) \quad (4)$$

where

$loss$ is the function to get classification error. For exsample, cross entropy function or mean squared error.

$w_f$ is the weighting coefficient for each $channel$, which is also learned by MF_DNET.

The following is the detail of training and predicting algorithm.

Algorithm 1 Training algorithm

```
1 Pick some frequencies in specified range
2 for each frequency
      Get the frequency-channel at each layer
      At output plane, get modulus for each frequency-channel
3 Sum up all output modulus with the weighting coefficient
4 Get loss by the formula (4)
5 Get gradient from the loss function and call back-propagation to improve the model
```

Algorithm 2 Predicting algorithm for input image

> 1 Load trained model for each frequency
> 2 For each model
>     Get the frequency-channel at each layer
>     At output plane, get modulus for each frequency-channel
> 3 Sum up all output modulus with the weighting coefficient
> 4 Call cross-entropy classification function to get output vector
> 5 The class number is the index corresponding to maximum value

As figure 1 showes, there are some modifications to implement MF_DNet. At the input layer, we transmit different optical pulse one by one. The output layer need an extra optical accumulator to merge the signals at different channels. MF_DNet still retains the advantage of DNNet. That is, MF_DNet would get predictions at the speed of light. It's pure passive architecture, no additional power consumption.

## 3. Results and discussion

We tested MF_DNet on two MNIST-like datasets(fasion-MNIST, EMNIST[19]), which is harder than MNIST. MNIST is the simplest testing datasets in image classification problems. It's so trivial that nearly no machine learning teams will use MNIST as testing dataset. Since DNNet has good accuary of in MNIST(higher than 90%) [1] , we should test it by more complex datasets. fasion-MNIST consists of a training set consisting of 60000 examples belonging to 10 different classes and a test set of 10000 examples. Figure 2 showes some pictures of fasion-MNIST. EMNIST is a set of handwritten character digits derived from the NIST Special Database 19. We use the "balanced" split of EMNIST which has 47 classes. Figure 3 showes some pictures of EMNIST.

Figure 4 is the learning curves on fasion-MNIST. Figure 5 is the learning curves on EMNIST. It's clear that with more frequency channels, MF_DNet would get much higher accuracy than standard DDNet. In the case fasion-MNIST, the accuracy increased from 73% to 85%. In the case EMNIST, the accuracy increased from 50% to 73%.

In the case of EMNIST, we must admit the performance on the EMNIST is poor (only 73%). But this is not the question. Just as said in the start of this article, diffractive is still in its infancy. With more structures and algorithms, we are sure there will be a great improvement. Just like the RELU, batch normalization, drop out…, these algorithms improve the accuracy of deep CNN drastically.

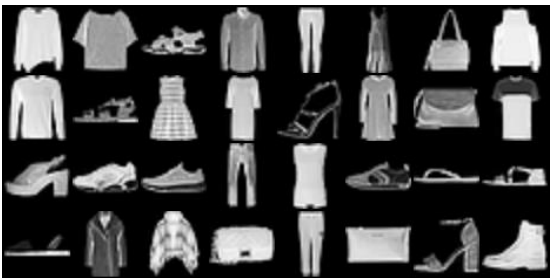
Figure 2 fasion-MNIST datasets (10 classes)

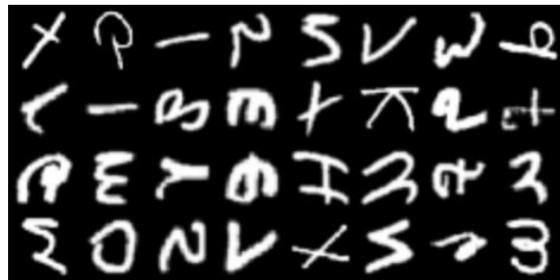
Figure 3 EMNIST datasets(47 classes)

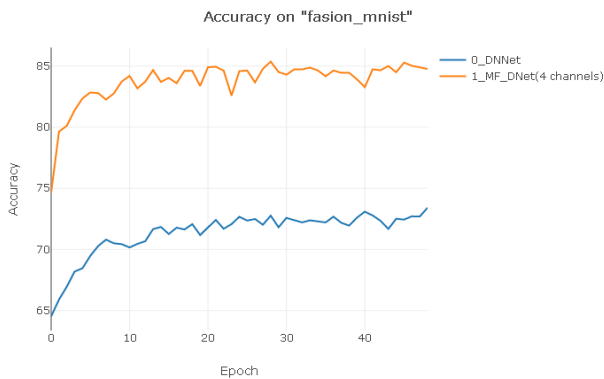
Figure 4 The learning cuvers on fasion-MNIST

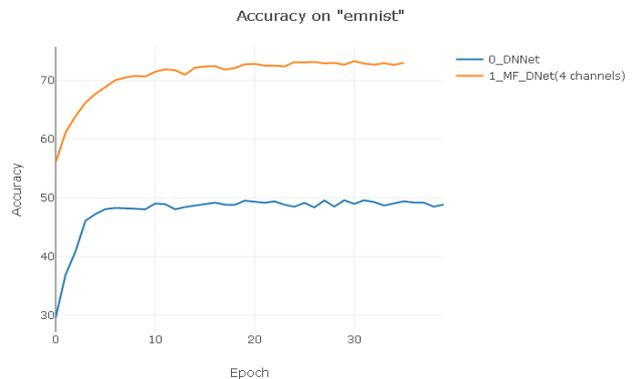
Figure 5 The learning cuvers on EMNIST

# 4 Conclusions

We give some in-depth analysis of DNNet's structure. We clarify that modulation process in DNNet is actually optical activation function. We improve the accuracy of DNNet with multiple frequency-channels. We released a new open source package ONNet. The main object of ONNet is to help people to simulation and study of optical neural networks. As verified by some datasets, the learning ability of DNNet is still weak. We need more structures and algorithms to improve its ability. Multiple frequency-channels is a effective algorithm for this task. And ONNet is a high performance and user-friendly package for more study and trials.